\documentclass[letterpaper]{article} 
\usepackage{aaai2027}    
\usepackage[hyphens]{url}            
\usepackage{graphicx}                
\usepackage{natbib}                  
\usepackage{caption}                 
\usepackage{subcaption}
\definecolor{codepink}{rgb}{1, 0.38, 0.45}
\usepackage{booktabs}
\usepackage{amsmath}
\usepackage{amssymb}
\usepackage{multirow}
\usepackage{xcolor} 
\nocopyright
\newcommand{\ibkd}{iBKD}
\newcommand{\ibam}{IBAM}
\newcommand{\Fcnn}{F^{\mathrm{cnn}}}
\newcommand{\Fvt}{F^{\mathrm{vt}}}

\newcommand{\Ltask}{\mathcal{L}_{\mathrm{task}}}
\newcommand{\Lalign}{\mathcal{L}_{\mathrm{align}}}
\newcommand{\Lfuse}{\mathcal{L}_{\mathrm{fuse}}}
\newcommand{\Ltotal}{\mathcal{L}_{\mathrm{total}}}

\title{Grid-Preserving Knowledge Distillation:\\Transferring Convolutional Inductive Bias to Vision Transformers \\under Data Scarcity}
\author {
    Junyong Choi\textsuperscript{\rm 2},
    Cheolhyeon Park\textsuperscript{\rm 1},
    Jaehoon Cho\textsuperscript{\rm 1}\corresponding
}
\affiliations {
    \textsuperscript{\rm 1}School of Electronics and Avionics Engineering, Korea Aerospace University,
    \textsuperscript{\rm 2}Hyundai Motor Company\\
}

\begin{document}
\maketitle

\begin{abstract}
Vision Transformers demonstrate remarkable global modeling capacity but often underperform in data-scarce regimes. 
Distilling convolutional inductive biases from a CNN teacher provides an effective remedy while leaving the deployed model unchanged. However, general-purpose feature distillation transfers little in this setting. In CNN-to-CNN distillation, pooling, flattening, and logit-space projections remove the spatial grid that encodes locality and translation equivariance. Unlike a convolutional student, a ViT cannot readily reconstruct this structure on its own. In this paper, we propose \ibkd{}, a distillation framework that preserves the spatial grid throughout the entire transfer process. Its core module, the Inductive Bias Attention Module, aggregates features from all student layers onto the teacher's grid using learned weights. It then enhances structural cues through channel and deformable spatial attention and injects them via convolutional cross-attention operating directly between spatial grids rather than token sets. The module is used only during training, leaving the deployed model as an unmodified ViT with no inference overhead. Across seven Transformer backbones and six data-scarce benchmarks, \ibkd{} consistently outperforms both locality-guidance methods and general knowledge distillation baselines, with its advantage increasing as the amount of training data decreases.

\end{abstract}

\noindent\textbf{Code} --- {\color{codepink}\url{https://github.com/kau-aimslab/iBKD}}
\section{Introduction}
\label{sec:intro}


\begin{figure}[t]
\centering
\includegraphics[width=1\columnwidth]{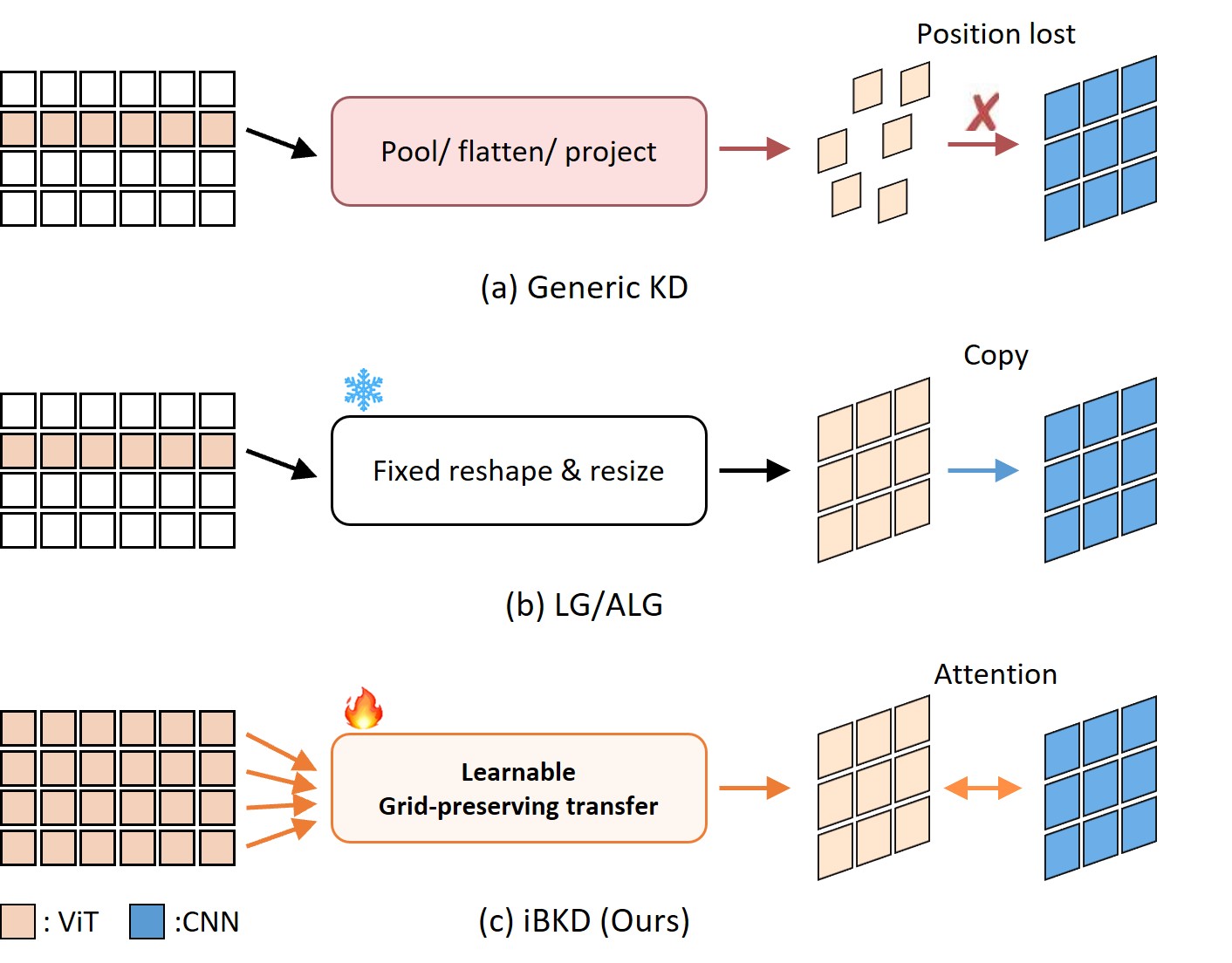}
\vspace{-8pt}
\caption{\textbf{Three transfer regimes.} {(a)} Generic KD pools or flattens features, losing positional correspondence. {(b)} LG/ALG keep the grid but fix the layer and target shape in advance. {(c)} \ibkd{} keeps the grid and learns the transfer, leveraging all layers and reading the CNN by attention.}
\label{fig:teaser}
\vspace{-13pt}
\end{figure}

Vision Transformers (ViTs) model long-range dependencies through global self-attention \citep{dosovitskiy2021image,liu2021swin}. However, unlike convolutional networks, they lack architectural inductive biases such as locality, translation equivariance, and hierarchical composition. These architectural differences lead ViTs and CNNs to learn visibly different representations. ViTs rely on early global attention, whereas CNNs gradually build up spatial structure \citep{raghu2021vision}. Consequently, ViTs exhibit a strong dependence on the amount of training data. They can surpass CNNs with large-scale pretraining, yet fall clearly behind when trained from scratch on small datasets \citep{dosovitskiy2021image,touvron2021training,liu2021efficient,li2022locality}. This data dependence may constrain their applicability in domains where data collection is expensive, such as medical imaging \citep{zhu2021hard} and fine-grained recognition \citep{nilsback2008automated}.

Existing remedies fall into three groups. Hybrid architectures incorporate convolutions into the Transformer \citep{wu2021cvt,dascoli2021convit,dai2021coatnet}. Self-supervised auxiliary objectives encourage spatially aware representations \citep{liu2021efficient}. CNN-guided knowledge distillation (KD) supervises intermediate ViT features using a convolutional teacher \citep{touvron2021training,li2022locality,rostand2025adaptive}. CNN-guided KD offers a practical advantage, as the teacher and auxiliary components are discarded after training. The resulting model is a vanilla ViT with zero inference overhead at deployment. Locality Guidance (LG) \citep{li2022locality} and Adaptive Locality Guidance (ALG) \citep{rostand2025adaptive} build on this approach through direct feature matching and report the strongest results in the data-scarce regime (Figure~\ref{fig:teaser}b).

It remains unclear why data-scarce CNN$\rightarrow$ViT transfer requires specialized methods at all, given that feature distillation is a mature field with strong general-purpose techniques \citep{tian2020contrastive,chen2021distilling,yang2022masked,hao2023one,gou2021knowledge}. We identify a structural reason. These methods were primarily designed for CNN$\rightarrow$CNN transfer. In this setting, the student's convolutions re-impose locality and equivariance on the transferred features. Transfer operators based on pooling, flattening, or logit-space projection \citep{tian2020contrastive,hao2023one} discard spatial structure during transfer. A CNN student can recover the corresponding locality and equivariance through its convolutional architecture. However, a ViT student has no such mechanism. The transfer path therefore needs to retain the spatial structure encoded by the \emph{grid}. Transfer operators that discard positional correspondence destroy exactly the structure the ViT student needs (Figure~\ref{fig:teaser}a). We test this explanation using a matched benchmark of generic KD methods, an operator-substitution ablation, and a grid-permutation experiment. In this experiment, we destroy positional correspondence while keeping the operators, parameters, and information content unchanged. This causes the transfer to collapse (in the Experiments section).

In this work, we build on this analysis and propose \textbf{Inductive Bias Knowledge Distillation (\ibkd{})}. Its core module, the \textbf{Inductive Bias Attention Module (\ibam{})}, preserves the spatial grid through three stages (Figure~\ref{fig:teaser}c). Learnable cross-architecture alignment aggregates multi-layer ViT tokens with learned weights and maps them back to the teacher's grid, replacing the fixed layer pairing used in prior work. Deformable enhancement amplifies structural cues along object boundaries. Convolutional cross-attention fusion computes attention directly between grids using $1{\times}1$ convolutional Q/K/V projections, rather than operating on flattened token sets. This grid-space design accounts for a 2.62-point accuracy difference. Each stage uses established operators by design. This choice allows us to attribute the measured gains to the grid-preserving principle rather than to newly introduced operators. \ibam{} is used only during training.

We further examine the source of the improvements through two controlled experiments. First, we train \ibam{} without the teacher. The module fails to recover vanilla accuracy, indicating that it provides no useful transferable signal on its own. Second, we retain both the module and the teacher while permuting the teacher grid. This intervention removes the gain entirely. Together, these results suggest that positional correspondence, rather than the operators or added capacity, is responsible for carrying the transferable bias. With the teacher and an intact grid, the three stages each contribute measurably to the final performance.

Our contributions are as follows:
\begin{itemize}
    \item We identify why generic feature distillation transfers little in the CNN$\rightarrow$ViT pairing. Transfer operators inherited from the CNN$\rightarrow$CNN regime discard the positional correspondence in which convolutional inductive biases are encoded, and a ViT student cannot restore it. We support this with a matched-protocol benchmark of general KD methods (CRD, ReviewKD, MGD, OFA) in this regime.
    \item We propose \ibkd{}/\ibam{}, a grid-preserving transfer mechanism combining learnable cross-architecture alignment, deformable enhancement, and convolutional cross-attention fusion, used only during training and adding zero inference overhead.
    \item We attribute the gains with three single-factor experiments. Removing the teacher prevents convergence, substituting token-space attention lowers accuracy, and permuting the teacher grid collapses the transfer, together showing that positional correspondence is what carries the transferred bias. \ibkd{} improves the state of the art on six benchmarks and seven Transformer backbones.
\end{itemize}

\begin{figure*}[t]
\centering
\includegraphics[width=2.1\columnwidth]{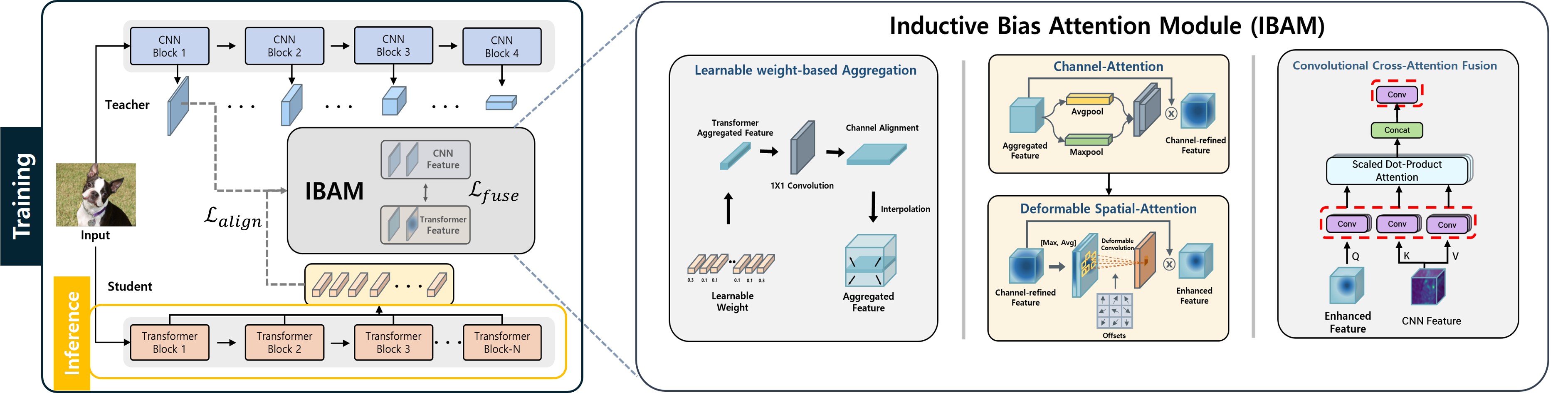}
\vspace{-3pt}
\caption{\textbf{Overview of \ibkd{}.} \emph{(left)} Training pipeline: the CNN teacher and \ibam{} supervise the ViT student through $\Lalign$ and $\Lfuse$, and are discarded entirely at inference, leaving an unmodified vanilla ViT. \emph{(right)} \ibam{} in detail: (a) alignment aggregates multi-layer student tokens and restores them onto the teacher's grid; (2) enhancement applies channel and deformable spatial attention on the grid; (3) fusion injects teacher priors via convolutional cross-attention with $1{\times}1$ Q/K/V projections.}
\label{fig:framework}\vspace{-8pt}
\end{figure*}

\section{Related Work}
\label{sec:related}

\subsection{Inductive Bias for Vision Transformers}
One line of work modifies the architecture itself. T2T-ViT \citep{yuan2021tokens} restructures tokenization to retain local structure, CvT \citep{wu2021cvt}, ConViT \citep{dascoli2021convit}, CoAtNet \citep{dai2021coatnet}, and NesT \citep{zhang2022nested} interleave convolution with attention, and LeViT \citep{graham2021levit}, MobileViT \citep{mehta2021mobilevit}, and LocalViT \citep{li2021localvit} add lightweight convolutional components. These designs permanently alter the deployed network. A second line transfers the bias during training only. DeiT \citep{touvron2021training} distills through an auxiliary token, DearKD \citep{chen2022dearkd} injects convolutional knowledge in an early training phase, Co-advise \citep{ren2022coadvise} pairs teachers with different inductive biases, LG \citep{li2022locality} supervises intermediate features with a fixed CNN teacher, and ALG \citep{rostand2025adaptive} schedules this supervision adaptively. Our work belongs to the second line, but differs in making explicit \emph{what} must survive the transfer, namely the spatial grid, and in learning the transfer end to end rather than fixing it a priori.

\subsection{Feature Distillation across Architectures}

Feature-level KD transfers intermediate representations from teacher to student \citep{hinton2015distilling,romero2014fitnets,gou2021knowledge}. We organize existing methods by their transfer operator, since under our analysis this determines whether spatial structure survives. Pooled or logit-space operators \citep{tian2020contrastive,zhao2022decoupled,hao2023one} discard the feature grid, while convolutional projectors \citep{chen2021distilling,yang2022masked}, direct feature matching \citep{li2022locality,rostand2025adaptive} and learned cross-architecture projection \citep{liu2022cross,lee2025customkd} retain it to varying degrees. In the CNN$\rightarrow$CNN setting these choices are largely interchangeable, because the student's convolutions restore spatial structure regardless of how the signal arrives. The CNN$\rightarrow$ViT setting removes this safety net, which is the regime studied here. 
A related direction is DearKD \citep{chen2022dearkd}, which distills convolutional knowledge into a ViT but focuses on \emph{when} the knowledge is injected through a phased schedule rather than on whether the transfer preserves the grid.
We instead make grid preservation an explicit principle, verify its causal role, and replace static matching with a learnable transfer.

\section{Method}
\label{sec:method}

\subsection{Preliminaries}
\label{sec:prelim}

Given an input $x$, a CNN teacher and a ViT student produce intermediate features $f_T(x)$ and $f_S(x)$. The teacher feature is $\Fcnn \in \mathbb{R}^{H_c \times W_c \times C_c}$, and student tokens are reshaped onto their native grid. Feature distillation trains the student with $\Ltask + \lambda\,\mathcal{L}_{\mathrm{FD}}$, where $\mathcal{L}_{\mathrm{FD}}$ penalizes a distance between (transformed) teacher and student features.

\subsection{Grid Collapse in Cross-Architecture Transfer}
\label{sec:setup}
Feature distillation matches values, but the teacher's inductive bias does not reside in the values. It resides in how they are arranged on the spatial grid. Locality is a statement about which positions are neighbors, and translation equivariance about how responses move when the input shifts, $f_T(\tau_\delta x) = \tau_\delta f_T(x)$. Both are properties of position, so once positional correspondence is discarded, neither can be expressed however accurately the values are matched.

We call a transfer operator \emph{grid-preserving} if its output is defined on the teacher grid and its value at position $p$ depends only on a fixed local neighborhood of $p$. Per-position ($1{\times}1$) convolutions, resolution-preserving resampling, and attention between grids satisfy this, whereas global pooling, flattening, and logit-space projection do not, since their output is no longer indexed by the grid. Only a grid-preserving target can express translation equivariance, because the constraint $T(f_T(\tau_\delta x)) = \tau_\delta\,T(f_T(x))$ applies the shift to the operator output and is undefined once that output leaves the grid. Preserving the grid is therefore necessary, though not sufficient, since whether the student acquires the bias remains a matter of optimization, which is why a preserved grid benefits from being paired with a learned transfer. In CNN$\rightarrow$CNN distillation this is immaterial, because the student's own convolutions re-impose the bias regardless of how the signal arrives. For a ViT student the transfer path is the only carrier, and the distinction becomes decisive. We verify this account in the Experiments section by benchmarking generic KD methods under an identical CNN$\rightarrow$ViT protocol, by substituting a single grid-space stage with its token-space counterpart, and by permuting the teacher grid.

\subsection{Inductive Bias Attention Module}
\label{sec:ibam}

\ibam{} instantiates grid-preserving transfer in three stages (Figure~\ref{fig:framework}), each built from established operators so that measured gains are attributable to the principle rather than to operator novelty. It takes the teacher feature $\Fcnn$ and student features from all $N$ blocks, $\{\Fvt_1, \dots, \Fvt_N\}$.

\subsubsection{Learnable Cross-Architecture Alignment.}
Prior locality-guidance methods match manually selected layer pairs, but which student depths are compatible with a given CNN teacher varies across architectures and cannot be fixed in advance. We instead learn the correspondence. For each teacher stage $g$, we aggregate all student layers with a stage-specific set of learned convex weights,
\begin{equation}
\mathbf{F}^{\mathrm{agg}}_{g} = \sum_{i=1}^{N} \alpha_{g,i}\, \mathbf{F}^{\mathrm{vt}}_{i}, \qquad
\alpha_{g,i} = \frac{e^{w_{g,i}}}{\sum_{j} e^{w_{g,j}}},
\label{eq:align}
\end{equation}
so that each teacher stage draws from the student depths most compatible with it rather than from a single shared pooling. Each aggregate is passed through a $1{\times}1$ convolution to match the teacher channel width and resampled to a common grid. The weights $w_{g,i}$ are initialized uniformly, so the model discovers this depth-to-stage correspondence during training rather than committing to a choice made in advance. All operations act per position or through local resampling, so the student representation reaches the teacher grid as a grid. We verify this choice against four manual strategies in the Experiments section (Table~\ref{tab:alignment}).

\subsubsection{Deformable Enhancement.}
Alignment establishes correspondence but does not make structural cues salient, and a rigid receptive field cannot follow object boundaries as they change shape across images. 
We therefore refine the aligned features with dual attention,
\begin{equation}
\tilde{\mathbf{F}}_{g} = \mathcal{A}_s\big(\mathcal{A}_c(\mathbf{F}^{\mathrm{agg}}_{g})\big),
\label{eq:enhance}
\end{equation}
where $\mathcal{A}_c$ and $\mathcal{A}_s$ denote channel and spatial attention.

\noindent\textbf{Channel attention.} We model inter-channel dependencies from pooled statistics,
\begin{equation}
\mathcal{A}_c(F) = \sigma\!\big(\mathrm{MLP}(P_{avg}) + \mathrm{MLP}(P_{max})\big) \odot F,
\label{eq:chatt}
\end{equation}
where $\sigma$ is the sigmoid, $\odot$ is element-wise multiplication, and $P_{avg}, P_{max} \in \mathbb{R}^{1\times1\times C}$ are pooled descriptors. This emphasizes the channels that carry structural rather than purely semantic information, following the channel-attention design of CBAM \citep{woo2018cbam}. The rescaling is identical at every position, so it leaves the grid untouched.

\noindent\textbf{Spatial attention.} To sharpen spatial saliency while retaining geometric flexibility, we apply a modulated deformable convolution \citep{zhu2020deformable} to concatenated pooled maps,
\begin{equation}
\mathcal{A}_s(F) = \sigma\!\big(\mathrm{DConv}([P_{avg}; P_{max}])\big) \odot F.
\label{eq:spatt}
\end{equation}
Unlike a standard convolution, which samples a fixed square neighborhood, the deformable kernel learns offsets and can therefore attend along non-rigid object boundaries. This matters here because the student's global patches and the teacher's fine-grained local responses rarely align on a regular lattice. The sampling still takes place within the grid at learned offsets, so spatial correspondence is preserved while the receptive field adapts.

\subsubsection{Convolutional Cross-Attention Fusion.}
Standard cross-attention operates on flattened token sets, where the queries, keys, and values are produced after positional structure has been serialized, so the attention map relates unordered tokens rather than spatial positions. We instead compute attention between grids with $h{=}4$ parallel heads,
\begin{equation}
\begin{aligned}
\mathbf{F}^{\mathrm{fused}}_{g} &= \mathrm{Conv}_{1\times1}\!\big(\mathrm{Concat}(\mathrm{head}_{g,1}, \dots, \mathrm{head}_{g,h})\big),\\
\mathrm{head}_{g,m} &= \mathrm{softmax}\!\left(\frac{Q_{g,m} K_{g,m}^{\top}}{\sqrt{d}}\right) V_{g,m},\\
Q_{g,m} &= \mathrm{Conv}_{1\times1}^{(m)}(\tilde{\mathbf{F}}_{g}),\quad
K_{g,m}, V_{g,m} = \mathrm{Conv}_{1\times1}^{(m)}(\mathbf{F}^{\mathrm{cnn}}_{g})
\end{aligned}
\label{eq:fusion}
\end{equation}
where each head draws its queries from the enhanced student feature $\tilde{\mathbf{F}}_{g}$ and its keys and values from the corresponding teacher stage feature $\mathbf{F}^{\mathrm{cnn}}_{g}$, all through $1{\times}1$ convolutions. Because every projection is a $1{\times}1$ convolution, each query, key, and value stays anchored to its position on the $H_c{\times}W_c$ grid. The per-head outputs are concatenated along the channel dimension and passed through a final $1{\times}1$ convolution that mixes the heads while keeping every position in place, so the fused feature returns on the same grid it started from.
 
Two properties follow. First, every projection is applied identically at each position, so it shares parameters across space in the same way a convolution does and is itself translation equivariant, whereas a projection defined over a flattened sequence is free to treat each position differently. Second, the attention map relates spatial positions to spatial positions, so the student reads the teacher's structure where that structure actually is. The stage therefore injects convolutional priors without giving up the global receptive field of attention, which is what makes it usable inside a ViT rather than in place of one.

\begin{table*}[t]
\centering
\caption{\textbf{Comparison with locality-guidance methods.} Top-1 accuracy (\%) on CIFAR-100, Flowers-102, Chaoyang, and CUB-200 with a ResNet-56 teacher (70.43 / 66.33 / 77.20 / 36.40) across seven Transformer backbones. \textbf{Bold} indicates the best result.}
\label{tab:main}
\vspace{-5pt}
\small
\setlength{\tabcolsep}{3pt}
\begin{tabular}{l cccc cccc cccc cccc}
\toprule
& \multicolumn{4}{c}{CIFAR-100} & \multicolumn{4}{c}{Flowers-102} & \multicolumn{4}{c}{Chaoyang} & \multicolumn{4}{c}{CUB-200} \\
\cmidrule(lr){2-5}\cmidrule(lr){6-9}\cmidrule(lr){10-13}\cmidrule(lr){14-17}
Student & Van. & LG & ALG & Ours & Van. & LG & ALG & Ours & Van. & LG & ALG & Ours & Van. & LG & ALG & Ours \\
\midrule
DeiT-Ti & 65.08 & 77.38 & 81.98 & \textbf{82.42} & 50.06 & 67.02 & 68.54 & \textbf{70.31} & 82.00 & 83.26 & 83.50 & \textbf{86.35} & 17.69 & 44.51 & 47.70 & \textbf{48.36} \\
ConViT  & 74.87 & 76.35 & 81.84 & \textbf{82.16} & 57.45 & 65.31 & 68.02 & \textbf{71.33} & 80.93 & 82.00 & 83.03 & \textbf{85.04} & 22.94 & 45.55 & 51.05 & \textbf{53.18} \\
CvT     & 74.29 & 76.78 & 76.81 & \textbf{77.29} & 60.82 & 67.13 & 69.07 & \textbf{69.85} & 80.04 & 82.42 & 83.74 & \textbf{84.02} & 29.19 & 43.01 & 46.63 & \textbf{47.98} \\
PiT     & 73.16 & 76.98 & 77.67 & \textbf{78.19} & 56.12 & 66.78 & 68.00 & \textbf{68.31} & 81.53 & 83.45 & 83.69 & \textbf{84.48} & 20.16 & 42.37 & 43.63 & \textbf{44.44} \\
PvTv2   & 77.21 & 75.90 & 79.19 & \textbf{79.22} & 67.89 & 65.70 & 68.34 & \textbf{75.89} & 82.52 & 82.84 & 83.50 & \textbf{84.30} & 47.15 & \underline{45.43} & 50.05 & \textbf{52.73} \\
T2T-7   & 68.00 & 76.48 & 79.56 & \textbf{80.43} & 66.14 & 67.57 & 71.06 & \textbf{72.52} & 78.78 & 81.44 & 81.81 & \textbf{86.45} & 26.73 & 46.63 & 50.57 & \textbf{54.59} \\
T2T-14  & 69.93 & 78.18 & 80.75 & \textbf{81.62} & 64.95 & 71.15 & 71.30 & \textbf{74.11} & 75.04 & 82.00 & 83.50 & \textbf{86.21} & 19.90 & 46.05 & 48.45 & \textbf{49.00} \\
\bottomrule
\end{tabular}
\end{table*}

\begin{table*}[t]
\centering
\caption{\textbf{General knowledge distillation under the CNN to ViT protocol.} Top-1 accuracy (\%) with the identical ResNet-56 teacher and DeiT-Ti student. Values below the vanilla baseline are \underline{underlined}. 
\textbf{Bold} is best.} \vspace{-3pt}
\label{tab:kdbaselines}
\small
\setlength{\tabcolsep}{2.6pt}
\begin{tabular}{ll ccc}
\toprule
Method & Transfer operator & C-100 & Flowers & Chaoyang \\
\midrule
Vanilla DeiT-Ti & --- & 65.08 & 50.06 & 82.00 \\
\midrule
\multicolumn{5}{l}{\emph{Operators that discard the grid}} \\
KD \citep{hinton2015distilling}     & Logits             & 69.10 & \underline{48.95} & \underline{74.05} \\
CRD \citep{tian2020contrastive}     & Pooled contrastive & 68.59 & \underline{49.06} & \underline{79.85} \\
OFA \citep{hao2023one}              & Logit-space proj.  & 67.73 & \underline{46.41} & \underline{78.03} \\
\midrule
\multicolumn{5}{l}{\emph{Operators that partly retain the grid}} \\
ReviewKD \citep{chen2021distilling} & Conv.\ projector   & 75.65 & 61.88 & 82.75 \\
MGD \citep{yang2022masked}          & Conv.\ projector (masked) & 75.68 & 54.66 & \underline{81.81} \\
\midrule
\multicolumn{5}{l}{\emph{Operators that preserve the grid}} \\
LG \citep{li2022locality}           & Direct match (static)    & 77.38 & 67.02 & 83.26 \\
ALG \citep{rostand2025adaptive}     & Scheduled match (static) & 81.98 & 68.54 & 83.50 \\
\textbf{\ibkd{} (Ours)}             & \textbf{Grid-space, learnable} & \textbf{82.42} & \textbf{70.31} & \textbf{86.35} \\
\bottomrule
\end{tabular} 
\end{table*}
\subsection{Training Objective}
\label{sec:objective}

The student is trained with
\begin{equation}
\Ltotal = \Ltask + \beta_e\Big[\lambda \sum_{g=1}^{G} \Lfuse^{g} + (1-\lambda) \sum_{g=1}^{G} \Lalign^{g}\Big],
\label{eq:total}
\end{equation}
with $\Lfuse^{g} = \|\mathbf{F}^{\mathrm{fused}}_{g} - \mathbf{F}^{\mathrm{cnn}}_{g}\|_2^2$ and $\Lalign^{g} = \|\mathbf{F}^{\mathrm{agg}}_{g} - \mathbf{F}^{\mathrm{cnn}}_{g}\|_2^2$, where $G$ is the number of teacher stages, $\Ltask$ is the cross-entropy loss, and $\lambda = 0.25$ unless stated otherwise. The only term that changes during training is $\beta_e$, which varies the strength of the distillation signal across epochs following the adaptive schedule of \citet{rostand2025adaptive}. $\Lalign$ supervises spatial and semantic correspondence at the aligned stage, while $\Lfuse$ supervises the fused output. Both losses are positionwise, so the supervision itself requires the grid correspondence that \ibam{} maintains. At inference, the teacher and \ibam{} are discarded.

\section{Experiments}
\label{sec:experiments}

\textbf{Datasets.} We evaluate on six benchmarks. CIFAR-10/100 \citep{krizhevsky2009learning} contain 50{,}000 training and 10{,}000 test images each, Flowers-102 \citep{nilsback2008automated} contains 8{,}189 images across 102 categories, Chaoyang \citep{zhu2021hard} contains 4{,}021 training and 2{,}139 test colorectal histopathology patches with label noise, CUB-200 \citep{wah2011caltech} contains 5{,}994 training and 5{,}794 test images across 200 fine-grained bird species, and Tiny-ImageNet \citep{le2015tiny} contains 200 classes with 500 training images per class.

\textbf{Models.} Following \citet{li2022locality}, the teacher is ResNet-56 \citep{he2016deep} for CIFAR, Flowers, CUB, and Chaoyang, and ResNet-50 for Tiny-ImageNet with a $3{\times}3$ stem and no initial max-pooling. All teachers, including the Tiny-ImageNet ResNet-50, are trained from scratch on the target training split only. No external pretraining is used anywhere in the pipeline, so the setting is strictly data-scarce and measured gains reflect transferred structure rather than leaked large-scale knowledge. Students span DeiT-Ti \citep{touvron2021training}, T2T-7/14 \citep{yuan2021tokens}, PiT \citep{heo2021rethinking}, PvTv2 \citep{wang2022pvt}, CvT \citep{wu2021cvt}, and ConViT \citep{dascoli2021convit}.


\textbf{Implementation.} We follow the training protocol of LG and ALG \citep{li2022locality,rostand2025adaptive} so that all methods are compared under identical conditions. Every student trains with AdamW \citep{loshchilov2017decoupled} at an initial learning rate of $5{\times}10^{-4}$, weight decay $0.05$, a cosine schedule with label smoothing $0.1$, and mixed precision, at resolution $224{\times}224$, while the CNN teacher operates at $32{\times}32$ as in LG and ALG. The number of epochs, batch size, and warm-up match the LG and ALG setup for each dataset and are listed per dataset in Appendix, with a batch size of $128$ on CIFAR-100 and $64$ on Flowers-102 and Chaoyang. The deformable kernel is $5{\times}5$ (Appendix) and $\lambda = 0.25$. Unless noted otherwise, every reported number is the mean over three runs with seeds $\{1,2,3\}$, and we report the standard deviation alongside the main comparisons. All experiments run on a single NVIDIA H200 GPU.

\textbf{Baseline protocol.} For Table~\ref{tab:kdbaselines} we run author-released implementations of KD \citep{hinton2015distilling}, CRD \citep{tian2020contrastive}, ReviewKD \citep{chen2021distilling}, MGD \citep{yang2022masked}, and OFA \citep{hao2023one} on CIFAR-100, Flowers-102, and Chaoyang with the identical ResNet-56 teacher, DeiT-Ti student, schedule, and augmentation, searching hyperparameters separately for each method. Appendix reports the selected hyperparameters and the token-to-grid adaptation.

We choose baselines that span the range of transfer operators, from logit-space and pooled embeddings to convolutional projectors, since the operator is the variable our analysis turns on.
CRD represents pooled embeddings, ReviewKD and MGD represent convolutional projectors, and OFA represents logit-space projection, which together span the range from discarding the grid to partly retaining it. This is what lets Table~\ref{tab:kdbaselines} read as a single graded axis rather than as a list of unrelated systems.

\subsection{Comparison with Prior Methods}
\label{sec:sota}

\textbf{Locality-guidance methods.} Table~\ref{tab:main} reports results across seven backbones on CIFAR-100, Flowers-102, Chaoyang, and the fine-grained CUB-200. \ibkd{} outperforms LG and ALG on every backbone and dataset, so the mechanism is not tied to a particular Transformer family and applies whether or not the backbone already carries built-in inductive bias. Margins are largest for backbones without endogenous convolutional components and on fine-grained or domain-shifted data, where the transferred structure cannot be substituted by the student's own capacity.

\noindent\textbf{General distillation methods.} As our analysis predicts, the methods in Table~\ref{tab:kdbaselines} do not perform uniformly, and what each transfers tracks how much of the spatial grid its operator keeps. On CIFAR-100, methods that transfer through logits or pooled embeddings recover the least, gaining $2.65$ to $4.02$ points over the vanilla student. Convolutional projectors, which retain part of the spatial layout, recover roughly two and a half times as much, gaining $10.57$ and $10.60$. Grid-preserving methods recover the most, from $12.30$ for LG to $17.34$ for \ibkd{}. The three groups do not overlap, which is the ordering our account predicts.

The smaller datasets separate the groups more sharply. On Flowers-102, every method in the first group falls \emph{below} the vanilla student, by $1.00$ to $3.65$ points, while the convolutional projectors gain $4.60$ and $11.82$ and the grid-preserving methods gain $16.96$ to $20.25$. Here the sign of the effect, not only its size, depends on whether the grid survives. CIFAR-100 supplies 50{,}000 training images, enough for a ViT to acquire some spatial regularity on its own, so even a structure-free signal helps. Flowers-102 and Chaoyang are an order of magnitude smaller, where this self-recovery is unavailable and spatially incoherent supervision instead becomes harmful.

\noindent\textbf{A teacher that the student already outperforms.} Two datasets make the mechanism unusually clear. On Chaoyang the ResNet-56 teacher reaches $77.20$, which is $4.80$ points \emph{below} the vanilla DeiT-Ti student at $82.00$, yet \ibkd{} lifts the student to $86.35$, or $9.15$ points above its own teacher. CUB-200 is more extreme still. It is fine-grained, and because the CNN teacher operates at $32{\times}32$, much of the detail needed to separate species is lost, so the teacher reaches only $36.40$, far below every student. Even so, \ibkd{} raises the DeiT-Ti student from $17.69$ to $48.36$ and the T2T-7 student to $54.59$, roughly $18$ points above the teacher. Copying a weaker model's values or decisions cannot produce this on either dataset. What is being transferred is the way the teacher organizes space rather than what it predicts. The same reading explains why static matching alone can even harm a strong student, since LG falls below the vanilla PvTv2 on CUB-200, while the learnable transfer in \ibkd{} recovers and surpasses it, and why logit-based methods lose the most ground where the teacher is weakest.

\noindent\textbf{Scaling to a larger regime.} Prior locality-guidance methods were evaluated only on small datasets such as CIFAR-100, Flowers-102, and Chaoyang. Tiny-ImageNet is an order of magnitude larger with 200 classes, and thus tests whether grid-preserving transfer remains useful as the regime grows. On DeiT-Ti, \ibkd{} lowers Top-1 error to $29.11\%$, improving over ALG at $30.17\%$ and LG at $30.38\%$, and the advantage holds across all DeiT capacities (Table~\ref{tab:tinyin}).

\begin{table}[t]
\centering
\caption{\textbf{Tiny-ImageNet.} Top-1 / Top-5 error (\%) with a ResNet-50 teacher trained from scratch. \textbf{Bold} is best.}  \vspace{-5pt}
\label{tab:tinyin}
\small
\begin{tabular}{l cc cc cc}
\toprule
& \multicolumn{2}{c}{DeiT-Ti} & \multicolumn{2}{c}{DeiT-S} & \multicolumn{2}{c}{DeiT-B} \\
\cmidrule(lr){2-3}\cmidrule(lr){4-5}\cmidrule(lr){6-7}
Method & T-1 & T-5 & T-1 & T-5 & T-1 & T-5 \\
\midrule
Vanilla & 51.07 & 27.69 & 51.27 & 28.76 & 52.72 & 30.28 \\
LG      & 30.38 & 12.39 & 30.27 & 12.82 & 30.76 & 13.09 \\
ALG     & 30.17 & 12.14 & 28.88 & 12.30 & 29.60 & 12.80 \\
Ours    & \textbf{29.11} & \textbf{11.75} & \textbf{28.84} & \textbf{11.88} & \textbf{29.02} & \textbf{12.65} \\
\bottomrule
\end{tabular}  \vspace{-3pt}
\end{table}

\begin{table}[t]
\centering
\caption{\textbf{Module and loss ablation} (CIFAR-100, DeiT-Ti). SpAtt is C for a standard $5{\times}5$ convolution and D for a deformable convolution at matched receptive field. Attn is the space in which fusion attention is computed. $\mathcal{L}_a$ and $\mathcal{L}_f$ denote $\Lalign$ and $\Lfuse$.}\vspace{-5pt}
\label{tab:ablation}
\small
\setlength{\tabcolsep}{2.6pt}
\begin{tabular}{c ccccc cc c}
\toprule
\# & Align & ChAtt & SpAtt & Fuse & Attn & $\mathcal{L}_a$ & $\mathcal{L}_f$ & Top-1 \\
\midrule
1 & \checkmark & & & & --- & \checkmark & & 78.00 \\
2 & \checkmark & & & \checkmark & Grid & \checkmark & \checkmark & 81.40 \\
3 & \checkmark & \checkmark & & \checkmark & Grid & \checkmark & \checkmark & 81.77 \\
4 & \checkmark & & D & \checkmark & Grid & \checkmark & \checkmark & 81.97 \\
5 & \checkmark & \checkmark & D & \checkmark & Token & \checkmark & \checkmark & 79.80 \\
6 & \checkmark & \checkmark & C & \checkmark & Grid & \checkmark & \checkmark & 81.96 \\
7 & \checkmark & \checkmark & D & \checkmark & Grid & \checkmark & \checkmark & \textbf{82.42} \\
\bottomrule
\end{tabular}\vspace{-8pt}
\end{table}

\subsection{Attributing the Gains}
\label{sec:sourceofgain}

\begin{figure}[t]
\centering
\includegraphics[width=0.9\columnwidth]{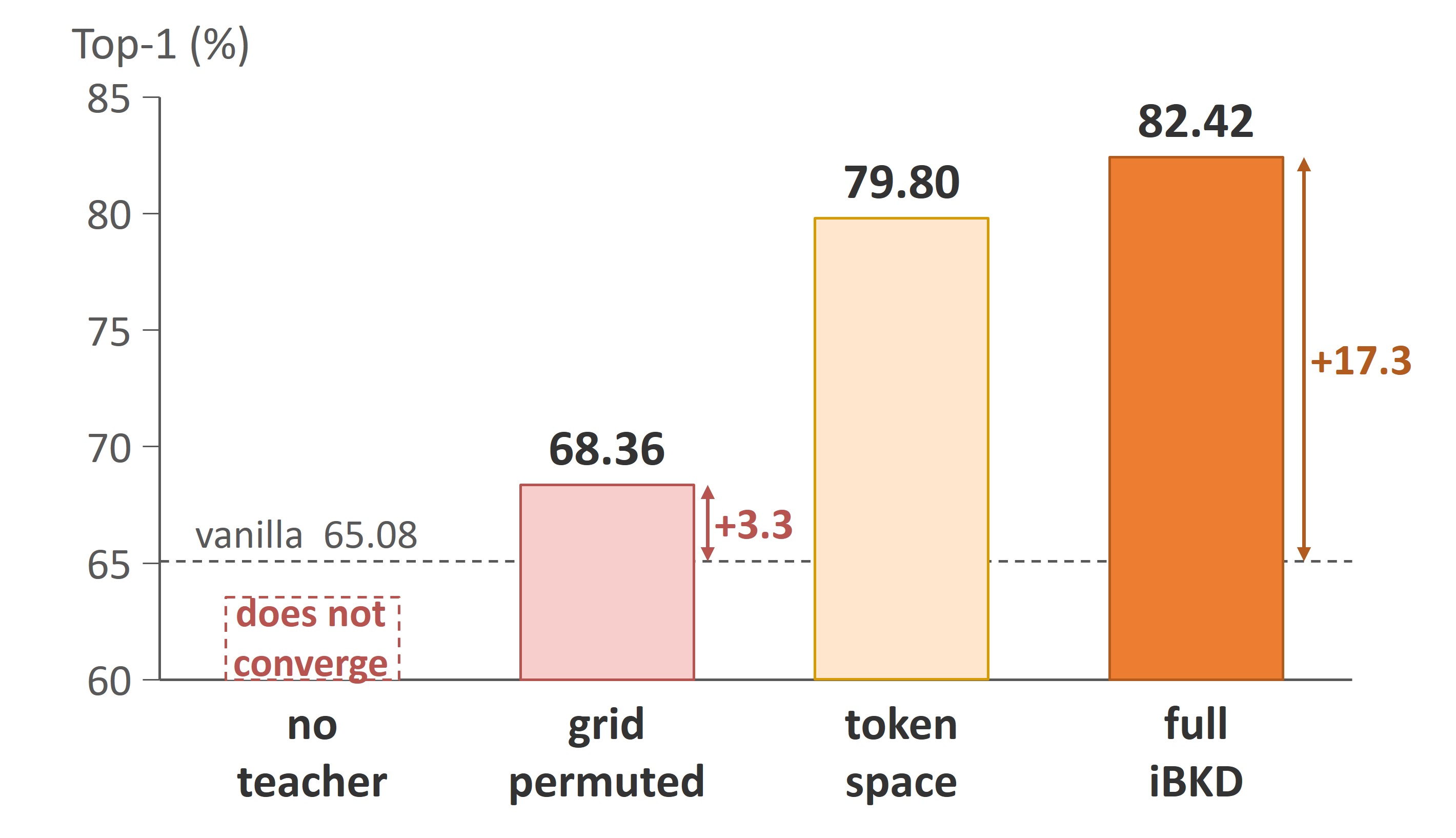}  \vspace{-8pt}
\caption{\textbf{Where the improvement comes from} (CIFAR-100, DeiT-Ti). Each configuration changes exactly one factor relative to full \ibkd{}.}
\label{fig:attribution}  \vspace{-8pt}
\end{figure}

Feature-level distillation raises an attribution question. Do the improvements come from the transferred knowledge, or from the parameters and operators that the auxiliary module adds? We separate these factors with three experiments, each changing exactly one element of the full framework and leaving everything else fixed (Figure~\ref{fig:attribution}).

\noindent\textbf{Removing the teacher.} We drop the CNN teacher and both distillation losses, replace $K$ and $V$ in Eq.~\eqref{eq:fusion} with self-features so that \ibam{} becomes an additional self-attention branch, and train with $\Ltask$ alone. Training fails to converge and does not reach vanilla accuracy. The module carries no useful prior of its own, and the gains originate in the teacher signal it transmits.

\noindent\textbf{Substituting token-space attention.} Replacing the grid-space fusion stage with a standard token-space cross-attention block, which computes fully connected Q, K, and V over flattened tokens with everything else unchanged, costs $2.62$ points and brings accuracy from $82.42$ to $79.80$. The same information, delivered without positional anchoring, transfers substantially less.

\noindent\textbf{Permuting the teacher grid.} The substitution above changes both the parameterization and the spatial behavior, so it cannot separate the operator from the grid. We therefore keep the convolutional cross-attention intact and apply one fixed random spatial permutation to the teacher features, using it consistently for $K$, $V$, and both positionwise loss targets. Operators, parameter counts, and information content are identical to full \ibkd{}, and only the positional correspondence is destroyed. Averaged over three permutation seeds, all of which converged normally, accuracy falls to $68.36$, which retains $3.28$ of the $17.34$ points that full \ibkd{} gains over the vanilla student. Roughly four fifths of the improvement depends on the teacher's features arriving in the right places rather than merely arriving. 

Two comparisons make this number concrete. Permuting the grid costs $11.44$ points more than substituting the operator, so positional correspondence matters considerably more than the form of the projection. And the permuted model lands at $68.36$, inside the $67.73$ to $69.10$ band occupied in Table~\ref{tab:kdbaselines} by KD, CRD, and OFA, the methods that discard the grid by design. Destroying the grid deliberately inside our framework reproduces the accuracy that those methods reach by construction, which links the graded pattern of Table~\ref{tab:kdbaselines} to a single causal factor rather than to differences among the methods themselves.

\begin{figure*}[t]
\centering
\includegraphics[width=2\columnwidth]{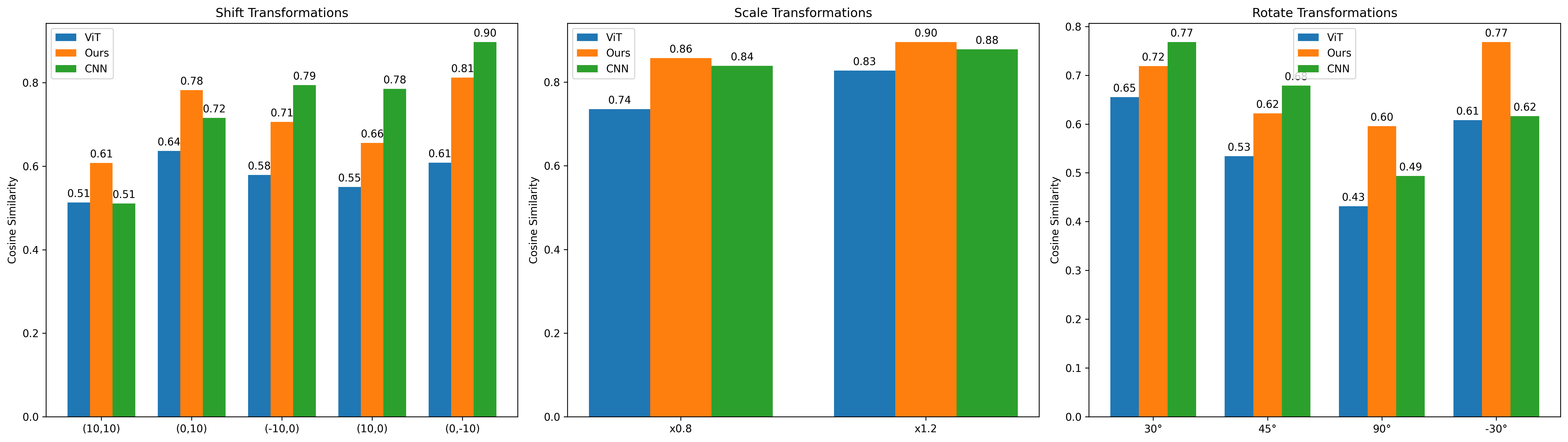} \vspace{-5pt}
\caption{\textbf{Feature consistency under geometric transformations.} Cosine similarity between features of the original and the transformed input, for shifts, scalings, and rotations. Higher is better. \ibkd{} students approach CNN-level stability, which indicates that translation equivariance itself was transferred rather than only accuracy.} 
\label{fig:geo}\vspace{-11pt}
\end{figure*}

\subsection{Ablation Studies}
\label{sec:ablation}

\textbf{Modules and losses.} Table~\ref{tab:ablation} ablates the \ibam{} stages, the spatial-attention convolution type, the fusion attention type, and the loss terms together. Alignment alone reaches $78.00$ in row 1. Adding grid-space fusion contributes $+3.40$ in row 2. Channel attention and deformable spatial attention add $+0.37$ and $+0.57$ individually in rows 3 and 4, and $+1.02$ jointly in row 7. Comparing rows 1 and 7 shows that the two losses together yield $+4.42$, and comparing rows 6 and 7 shows that deformable sampling adds $+0.46$ over a standard convolution at matched receptive field. Row 5 is the token-space substitution discussed above.

\noindent\textbf{Alignment strategy.} Table~\ref{tab:alignment} compares the learnable aggregation of Eq.~\eqref{eq:align} against four manual strategies, evaluated with $\Lalign$ only. The four manual strategies differ in how they pair student depths with teacher stages, yet they land within $0.49$ points of one another, between $76.21$ and $76.70$. Learning the correspondence instead gains $1.30$ over the best of them. The choice of manual rule matters little, and what matters is that the correspondence is learned at all, which is consistent with our claim that the compatible depth range cannot be fixed in advance.

\begin{table}[t]
\centering
\caption{\textbf{Alignment strategies} evaluated with $\Lalign$ only (CIFAR-100, DeiT-Ti). M denotes manual and L denotes learnable.}\vspace{-5pt}
\label{tab:alignment}
\small
\begin{tabular}{l c c}
\toprule
Strategy & M/L & Top-1 (\%) \\
\midrule
(1) Full sum of CNN features        & M & 76.21 \\
(2) Manual selection (LG-style)     & M & 76.50 \\
(3) Group-wise ($m$/$n$) matching   & M & 76.64 \\
(4) Stochastic selection            & M & 76.70 \\
(5) Learnable (ours)                & L & \textbf{78.00} \\
\bottomrule
\end{tabular}\vspace{-8pt}
\end{table}

\noindent\textbf{Loss balance.} Table~\ref{tab:lambda} sweeps $\lambda$ in Eq.~\eqref{eq:total}, which splits the distillation budget between the aligned stage and the fused output. Setting $\lambda = 0$ removes $\Lfuse$ and reproduces row 1 of Table~\ref{tab:ablation} at $78.00$. Once both terms are active, accuracy varies by only $1.10$ points across $\lambda \in [0.25, 1.0]$, so the balance is not a sensitive hyperparameter, with the aligned stage the more important term since $\lambda = 1$ is the weakest at $81.32$. We use $\lambda = 0.25$ throughout.

\begin{table}[t]
\centering
\caption{\textbf{Sensitivity to the loss balance $\lambda$} (CIFAR-100, DeiT-Ti). $\lambda = 0$ supervises only the aligned stage, which reproduces row 1 of Table~\ref{tab:ablation}, and $\lambda = 1$ supervises only the fused output.}\vspace{-8pt}
\label{tab:lambda}
\small
\setlength{\tabcolsep}{5pt}
\begin{tabular}{l ccccc}
\toprule
$\lambda$ & 0 & 0.25 & 0.5 & 0.75 & 1.0 \\
\midrule
Top-1 (\%) & 78.00 & \textbf{82.42} & 81.93 & 81.66 & 81.32 \\
\bottomrule
\end{tabular}\vspace{-8pt}
\end{table}

\subsection{Data Scarcity, Efficiency, and Bias Transfer}
\label{sec:scarcity}

\textbf{Extreme data scarcity.} Table~\ref{tab:scarcity} restricts CIFAR-10/100 to 20\%, 50\%, and 100\% of the training data. \ibkd{} leads in every regime, and its margin over ALG widens as data shrinks, reaching $+2.80$ at 20\% of CIFAR-100. Transferred structure matters most exactly where the student cannot learn it from data.

\begin{table}[t]
\centering
\caption{\textbf{Extreme data scarcity.} Top-1 accuracy (\%) with varying training-set fractions and a DeiT-Ti student.}\vspace{-8pt}
\label{tab:scarcity}
\small
\setlength{\tabcolsep}{4pt}
\begin{tabular}{l ccc ccc}
\toprule
& \multicolumn{3}{c}{CIFAR-10} & \multicolumn{3}{c}{CIFAR-100} \\
\cmidrule(lr){2-4}\cmidrule(lr){5-7}
Method & 20\% & 50\% & 100\% & 20\% & 50\% & 100\% \\
\midrule
Vanilla & 65.83 & 81.48 & 88.70 & 31.03 & 50.31 & 64.65 \\
LG      & 91.48 & 94.37 & 95.39 & 64.33 & 73.28 & 77.38 \\
ALG     & 91.39 & 95.18 & 96.29 & 64.82 & 74.90 & 81.13 \\
Ours    & \textbf{93.20} & \textbf{96.12} & \textbf{97.25} & \textbf{67.62} & \textbf{76.89} & \textbf{82.42} \\
\bottomrule
\end{tabular}\vspace{-8pt}
\end{table}

\noindent\textbf{Efficiency and convergence.} \ibkd{} adds no inference overhead, since the teacher and \ibam{} are used only during training and discarded at deployment, leaving an unmodified ViT with identical latency. On an epoch axis it also reaches any given accuracy in fewer epochs and converges higher (Figure~\ref{fig:curves}), which is what grid preservation should produce. 

\noindent\textbf{Does the bias itself transfer?} 
Accuracy alone does not establish that the student acquired CNN-like structure, so we measure the property directly. Under geometric perturbations, \ibkd{} students approach CNN-level stability under shift, scale, and rotation (Figure~\ref{fig:geo}).
Mean attention distance \citep{raghu2021vision} approaches that of ImageNet-pretrained models even though training used Tiny-ImageNet alone. Translation equivariance is one of the two properties we argued lives in the grid, and it is measurably higher in students trained with a grid-preserving transfer. Appendix reports the full analyses.

\section{Conclusion}
\label{sec:conclusion}

We asked why general feature distillation transfers so little from a CNN teacher to a ViT student, and traced the failure to transfer operators that discard the positional correspondence carrying convolutional inductive bias. This loss is harmless when the student can rebuild spatial structure with its own convolutions and destructive when it cannot. \ibkd{} keeps the grid intact along the whole transfer path, and three single-factor experiments show that the grid itself, rather than the added module or its operators, carries the transferred bias. Because grid preservation concerns spatial correspondence rather than a particular objective, we expect it to matter at least as much for dense prediction such as semantic segmentation and monocular depth estimation, where a prediction is made at every position. We leave this extension to future work.

\bibliography{aaai2027}

\clearpage

\appendix

\begin{figure*}
  \centering
  \begin{subfigure}[b]{0.24\textwidth}
    \centering\includegraphics[width=\linewidth]{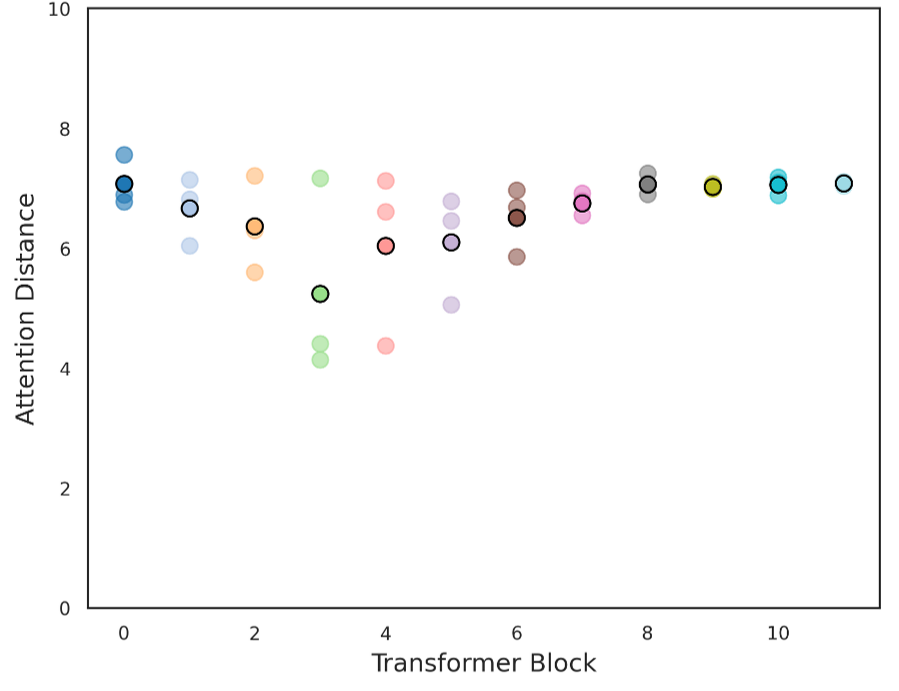}
    \caption{}
  \end{subfigure}\hfill
  \begin{subfigure}[b]{0.24\textwidth}
    \centering\includegraphics[width=\linewidth]{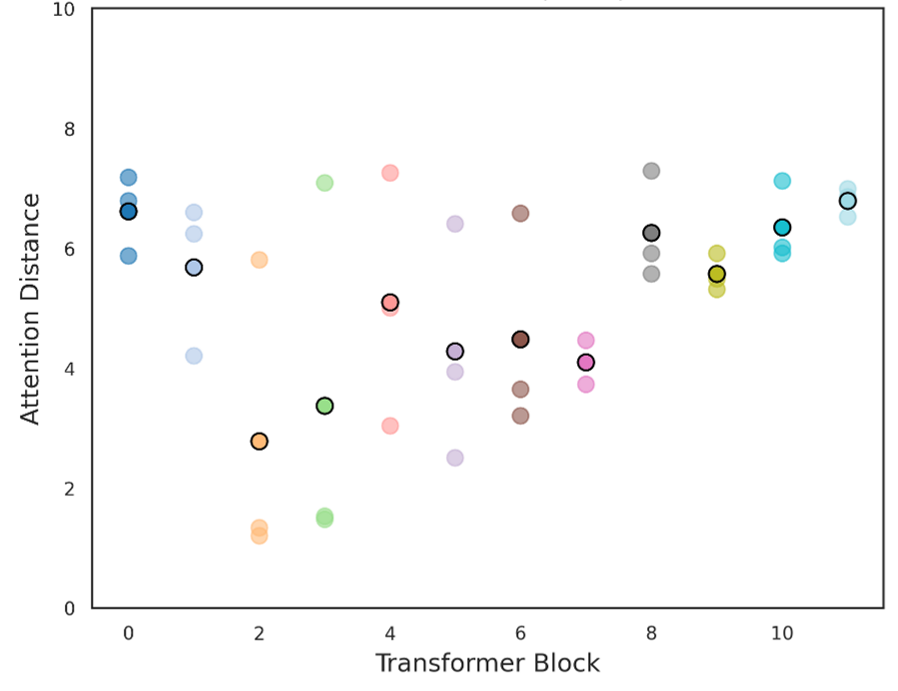}
    \caption{}
  \end{subfigure}\hfill
  \begin{subfigure}[b]{0.24\textwidth}
    \centering\includegraphics[width=\linewidth]{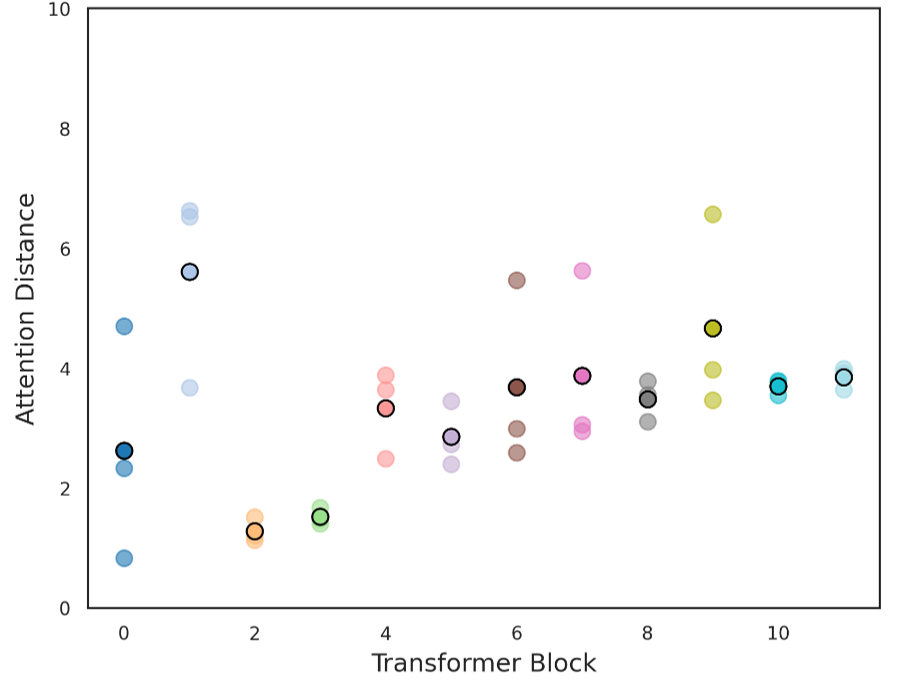}
    \caption{}
  \end{subfigure}\hfill
  \begin{subfigure}[b]{0.24\textwidth}
    \centering\includegraphics[width=\linewidth]{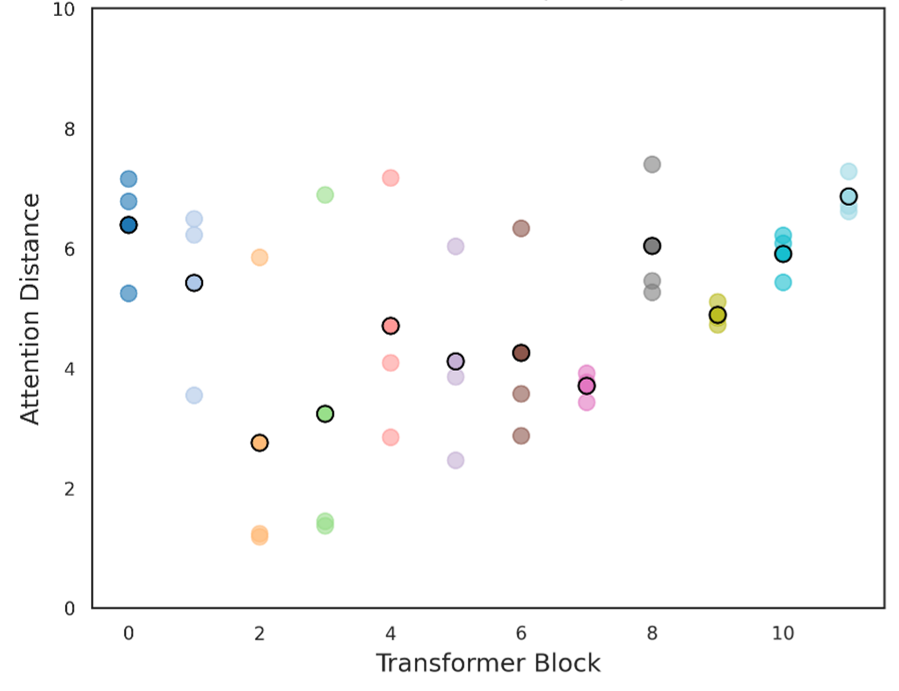}
    \caption{}
  \end{subfigure}
  \caption{Mean Attention Distance (MAD) across transformer layers. (a) DeiT-Ti (from scratch), (b) DeiT-Ti pretrained on ImageNet, (c) LG~\cite{li2022locality}, and (d) Ours. Our method achieves broad spatial attention comparable to ImageNet-pretrained models, despite training only on Tiny-ImageNet, while LG and vanilla DeiT remain more locally biased.}
  \label{fig:mad}
\end{figure*}

\section{Baseline Protocol}
\label{app:kdprotocol}
 
\textbf{Shared teacher and student.} All compared methods reuse a single fixed teacher per dataset rather than training a separate teacher each. The teacher is a CIFAR-style ResNet-56 ($6n{+}2$, $n{=}9$) trained at $32{\times}32$ resolution, following the recipe of \citet{li2022locality}. Teachers are trained with SGD (initial learning rate $0.1$, momentum $0.9$ with Nesterov, weight decay $5{\times}10^{-4}$ excluding bias and normalization parameters), a cosine schedule with no warm-up, and the official locality-guidance strong-augmentation path (random resized crop to $32$ with bicubic interpolation, horizontal flip, RandAugment \texttt{m9}, and random erasing with probability $0.25$). The CIFAR-100 teacher trains for $300$ epochs, the Flowers-102 teacher for $450$ epochs on the official train and validation split, and the Chaoyang teacher for $300$ epochs.
 
The student is a DeiT-Ti (\texttt{deit\_tiny\_patch16\_224}) trained from scratch at $224{\times}224$, with the teacher branch receiving the same student crop bilinearly resized to $32{\times}32$ so that crop and flip geometry are shared while each model keeps its own normalization. Every compared method uses AdamW with initial learning rate $5{\times}10^{-4}$, weight decay $0.05$, a cosine schedule, label smoothing $0.1$, and mixed precision. The per-dataset epochs, batch size, and warm-up are given in Table~\ref{tab:kdprotocol}, and only the method-specific transfer loss and its adapter differ across methods.

\begin{figure*}[t]
  \centering
  \begin{subfigure}[b]{0.16\linewidth}
    \centering\includegraphics[width=\linewidth]{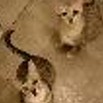}
    \caption{}
  \end{subfigure}\hfill
  \begin{subfigure}[b]{0.16\linewidth}
    \centering\includegraphics[width=\linewidth]{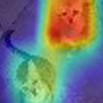}
    \caption{}
  \end{subfigure}\hfill
  \begin{subfigure}[b]{0.16\linewidth}
    \centering\includegraphics[width=\linewidth]{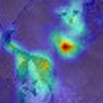}
    \caption{}
  \end{subfigure}\hfill
  \begin{subfigure}[b]{0.16\linewidth}
    \centering\includegraphics[width=\linewidth]{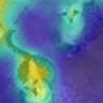}
    \caption{}
  \end{subfigure}\hfill
  \begin{subfigure}[b]{0.16\linewidth}
    \centering\includegraphics[width=\linewidth]{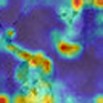}
    \caption{}
  \end{subfigure}\hfill
  \begin{subfigure}[b]{0.16\linewidth}
    \centering\includegraphics[width=\linewidth]{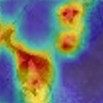}
    \caption{}
  \end{subfigure}
  \caption{Visualization of attention maps under low-data settings. (a) Input image, (b) ResNet~\cite{he2016deep}, (c) DeiT, (d) LG~\cite{li2022locality}, (e) ALG~\cite{rostand2025adaptive}, and (f) Ours. Our method achieves sharp and semantically aligned attention across multiple objects.}
  \label{fig:heatmap}
\end{figure*}

\begin{figure*}
  \centering
  \begin{subfigure}[b]{0.32\linewidth}
    \centering\includegraphics[width=\linewidth]{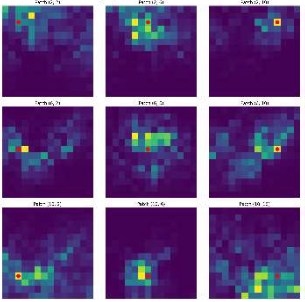}
  \end{subfigure}\hfill
  \begin{subfigure}[b]{0.32\linewidth}
    \centering\includegraphics[width=\linewidth]{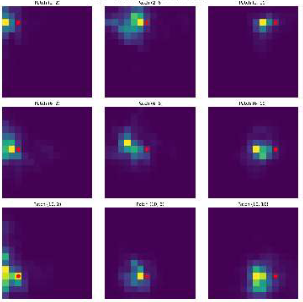}
  \end{subfigure}\hfill
  \begin{subfigure}[b]{0.32\linewidth}
    \centering\includegraphics[width=\linewidth]{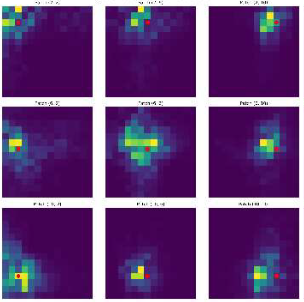}
  \end{subfigure}
  \vspace{2pt}
  \begin{subfigure}[b]{0.32\linewidth}
    \centering\includegraphics[width=\linewidth]{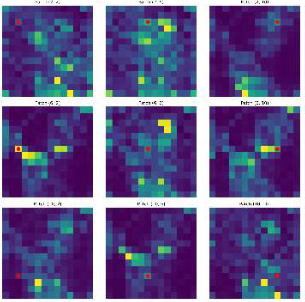}
  \end{subfigure}\hfill
  \begin{subfigure}[b]{0.32\linewidth}
    \centering\includegraphics[width=\linewidth]{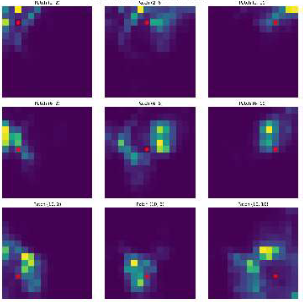}
  \end{subfigure}\hfill
  \begin{subfigure}[b]{0.32\linewidth}
    \centering\includegraphics[width=\linewidth]{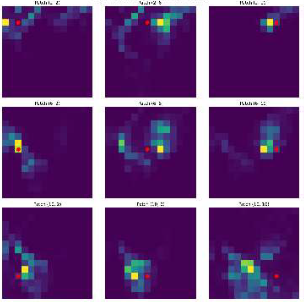}
  \end{subfigure}
  \vspace{2pt}
  \begin{subfigure}[b]{0.32\linewidth}
    \centering\includegraphics[width=\linewidth]{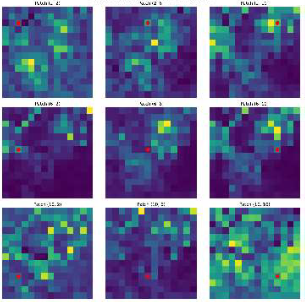}
    \caption{ViT}
  \end{subfigure}\hfill
  \begin{subfigure}[b]{0.32\linewidth}
    \centering\includegraphics[width=\linewidth]{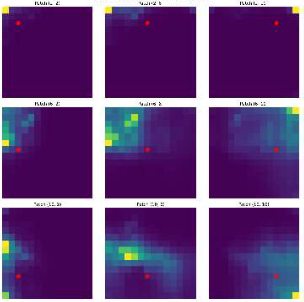}
    \caption{LG}
  \end{subfigure}\hfill
  \begin{subfigure}[b]{0.32\linewidth}
    \centering\includegraphics[width=\linewidth]{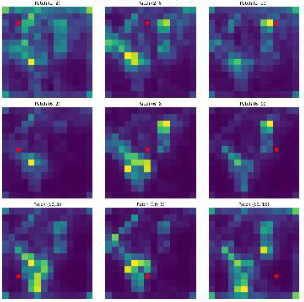}
    \caption{Ours}
  \end{subfigure}
  \caption{Visual comparison of attention maps between the baseline ViT, LG~\cite{li2022locality}, and our proposed \ibkd{}.}
  \label{fig:layerattn}
\end{figure*}

\begin{table}[h]
\centering
\caption{\textbf{Shared student protocol for Table~\ref{tab:kdbaselines}.} All KD baselines use these settings, differing only in the method-specific transfer loss and adapter.}
\label{tab:kdprotocol}
\small
\setlength{\tabcolsep}{3pt}
\begin{tabular}{l ccc}
\toprule
Setting & CIFAR-100 & Flowers-102 & Chaoyang \\
\midrule
Epochs      & 300 & 200 & 100 \\
Batch size  & 128 & 64  & 64 \\
Warm-up     & 20  & 5   & 5 \\
Optimizer   & \multicolumn{3}{c}{AdamW, lr $5{\times}10^{-4}$, wd $0.05$} \\
Schedule    & \multicolumn{3}{c}{cosine, label smoothing $0.1$, AMP} \\
Student res.\ / Teacher res. & \multicolumn{3}{c}{$224{\times}224$ / $32{\times}32$} \\
\bottomrule
\end{tabular}
\end{table}
 
\noindent\textbf{Method-specific settings and adapters.} Each method keeps the coefficients of its official implementation. KD uses a temperature of $4.0$ and a balance of $\alpha{=}0.9$ between the cross-entropy and logit terms, transferring only the class distribution so no spatial adapter is required. CRD contrasts a global-average-pooled ResNet stage-3 vector ($64$d) against the DeiT pre-logits ($192$d), both mapped to $128$d, with an NCE temperature of $0.07$ and memory momentum $0.5$, so no spatial correspondence is kept. ReviewKD fuses post-activation ResNet stages 1 to 3 (grids $32/16/8$, channels $16/32/64$) with a feature-loss weight of $0.6$ that ramps over $20$ epochs. MGD reconstructs masked features with weight $\alpha{=}7{\times}10^{-5}$ and mask probability $0.15$. OFA projects features into the logit space at temperature $1.0$ using student stages 1 to 4, deliberately abandoning the feature grid. The pattern that matters for our analysis is visible in these adapters, since KD, CRD, and OFA reduce the teacher to a vector or a logit while ReviewKD and MGD retain a convolutional feature grid.


\noindent\textbf{Kernel size.} Table~\ref{tab:kernel} varies the deformable kernel size across the seven backbones on Chaoyang. Kernel size $5$ gives the best or near-best accuracy on most backbones, for example $86.35$ on DeiT-Ti and $86.45$ on T2T-7, and we use it throughout. Smaller kernels tend to underperform on the deeper backbones, while larger kernels give no consistent gain, so a moderate receptive field balances local coverage and spatial selectivity.

\begin{table}[h]
\centering
\caption{\textbf{Deformable kernel size} (Top-1 accuracy, \%, Chaoyang) across the seven Transformer backbones. \textbf{Bold} is best in each column.}
\label{tab:kernel}
\small
\setlength{\tabcolsep}{3pt}
\begin{tabular}{l ccccccc}
\toprule
Kernel & DeiT-Ti & T2T-7 & T2T-14 & PiT & PvTv2 & ConViT & CvT \\
\midrule
$3{\times}3$ & 85.93 & 85.42 & 85.33 & \textbf{84.67} & 83.97 & 85.33 & \textbf{84.31} \\
$5{\times}5$  & \textbf{86.35} & \textbf{86.45} & \textbf{86.21} & 84.48 & \textbf{84.30} & 85.04 & 84.02 \\
$7{\times}7$ & 86.26 & 85.00 & 86.17 & 83.69 & 83.78 & \textbf{85.84} & 82.85 \\
\bottomrule
\end{tabular}
\end{table}

\section{Inductive Bias Analyses}
\label{app:bias}
This section expands the bias-transfer evidence summarized in Experiments section (Data Scarcity, Efficiency, and Bias Transfer). It collects the mean attention distance analysis, the layer-wise attention visualizations, the attention heatmaps, and the full geometric-consistency curves from which the numbers in the main text are drawn.
 
\noindent\textbf{Mean attention distance.} To quantitatively assess the spatial modeling behavior of ViTs under limited data, we measure the Mean Attention Distance (MAD)~\cite{raghu2021vision} across transformer layers. This metric reflects the average spatial extent a token attends to—higher values indicate broader, more global attention.

Figure~\ref{fig:mad} compares four models: (a) DeiT trained from scratch, (b) DeiT pretrained on ImageNet, (c) LG~\cite{li2022locality}, and (d) our iBKD. As shown in the Figure~\ref{fig:mad}(b) and (d), while the ImageNet-pretrained DeiT naturally achieves wide spatial coverage, iBKD—despite being trained only on Tiny-ImageNet—closely approximates this pattern. In contrast, vanilla DeiT and LG remain overly local or inconsistent, especially in early layers.
This result highlights a key strength of iBKD. It enables ViTs to acquire globally aware attention patterns comparable to those obtained via costly large-scale pretraining, by selectively injecting inductive bias from CNNs during training.

\begin{figure*}
  \centering
  \includegraphics[width=0.9\linewidth]{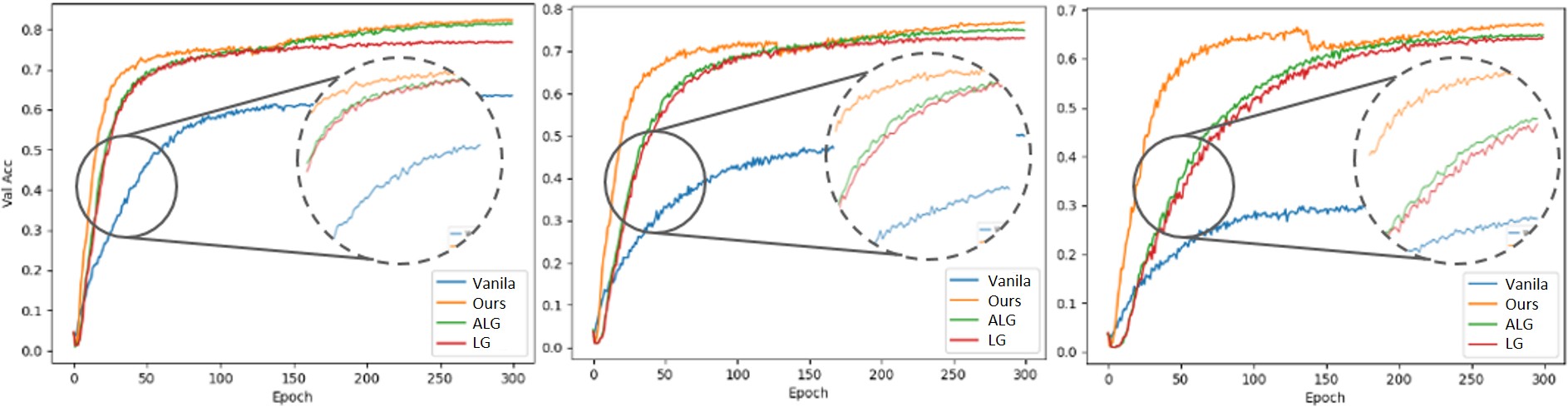}
  \caption{
  Validation accuracy curves on CIFAR-100 with different data scales. From left to right: Training convergence is visualized under 100\%, 50\%, and 20\% training data. Our method (orange line) consistently exhibits faster convergence and higher final accuracy than LG and ALG, especially under extreme data scarcity.}
  \label{fig:curves} \vspace{-8pt}
\end{figure*}

\noindent\textbf{Attention heatmaps.} We also compare attention heatmaps for a given input image from Tiny-ImageNet. As shown in Figure~\ref{fig:heatmap}, ResNet shows sharp localization due to strong inductive bias, while DeiT-Ti exhibits sparse focus. LG~\cite{li2022locality} and ALG~\cite{rostand2025adaptive} improve spatial awareness, but still suffer from diffuse or noisy attention. In contrast, our iBKD produces sharp and complete attention coverage on both salient regions (e.g., two cats), demonstrating that the proposed cross-attention distillation improves spatial generalization even in data-scarce regimes.

\noindent\textbf{Layer-wise attention.} In Figure~\ref{fig:layerattn}, we visualize the spatial attention patterns across layers by adopting the patch-based interpretability framework of Chefer et al.~\cite{chefer2021transformer}. A single query token (marked as a red dot) is selected, and its attention weights to all other patches are shown as heatmaps. The input image used is shown in Figure~\ref{fig:heatmap} (a).
As shown in Figure~\ref{fig:layerattn} (a), DeiT-Ti exhibits narrow and noisy activations, while (b) LG improves locality but often lacks global coherence. In contrast, (c) our iBKD highlights semantically consistent and spatially distributed regions from early layers, showing better integration of local structure and global reasoning.

\section{Additional Implementation Details}
\label{app:impl}
We give the details omitted from Experiments section. These include the exact teacher configurations for each dataset, the grid-permutation procedure used in Experiments section (Permuting the teacher grid).

\noindent\textbf{Grid-permutation procedure.} A single spatial permutation of the teacher grid is drawn once per run and held fixed for all iterations. It is applied consistently to $K$, to $V$, and to both positionwise loss targets, so that the information reaching the student is identical to the unpermuted model and only the spatial correspondence changes.

\noindent\textbf{Computing environment.} All experiments run on a single NVIDIA H200 GPU with Python 3.10.12, PyTorch 2.11.0 (CUDA 13.0), torchvision 0.26.0, and timm 1.0.27. For the general knowledge distillation comparison in Table~\ref{tab:kdbaselines}, we do not reimplement the student but load the DeiT-Tiny model \texttt{deit\_tiny\_patch16\_224} from timm with \texttt{pretrained=False}, so that it is trained from scratch without any pretrained weights. For the seven backbones in the CUB-200 results of Table~\ref{tab:main}, namely DeiT-Ti, ConViT-Ti, CvT-13, PiT-Ti, PVTv2-B0, T2T-ViT-7, and T2T-ViT-14, we use the model code and configuration files from the official LG tiny-transformers repository, and these backbones are likewise trained from scratch without pretraining.

\section{Convergence Curves}
\label{app:curves}
Figure~\ref{fig:curves} plots validation accuracy against training epochs for LG, ALG, and \ibkd{} at 100, 50, and 20 percent of CIFAR-100. On an epoch axis, \ibkd{} reaches any given accuracy in fewer epochs and converges higher, which is the basis for the convergence statement in the Experiments section. Per-epoch wall-clock time is higher for \ibkd{}, so the comparison is drawn on epochs rather than time.

\begin{figure*}
    \centering
    \includegraphics[width=0.5\textwidth]{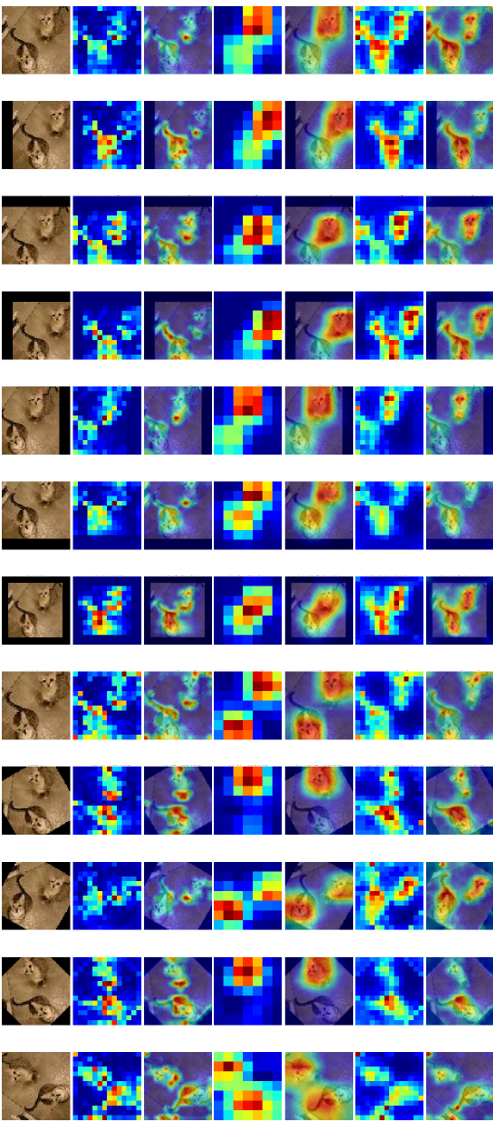}
    \caption{\textbf{Comparison of CAM and attention maps across architectures.} From left to right: input image, ViT (DeiT-Ti) CAM, ViT (DeiT-Ti) attention, CNN (ResNet) CAM, CNN attention, ours (iBKD) CAM, and ours attention. Our iBKD generates more spatially coherent and semantically complete activations, capturing both local structures and global context even under limited data.}
    \label{fig:cat}
\end{figure*}

\subsection*{Qualitative Analysis of Geometric Robustness}
Figure~\ref{fig:cat} compares Class Activation Maps and attention maps across architectures under translation, scaling, and rotation. We report these as qualitative observations on a set of representative inputs rather than as quantitative evidence, which Figure~\ref{fig:geo} provides.

\noindent\textbf{Comparison with the vanilla ViT.} On the examples shown, the vanilla DeiT-Ti often produces scattered activations when data is limited, while \ibkd{} tends to concentrate on the object region. This is consistent with the transferred inductive bias helping the tokens organize spatial information.

\noindent\textbf{Comparison with the CNN teacher.} The CNN teacher shows strong local responses but its activations often do not cover the whole object, whereas \ibkd{} attends more broadly while remaining on the object, which is consistent with combining convolutional locality and attention-based global context.

\noindent\textbf{Robustness to geometric variations.} Under rotation and scaling, \ibkd{} activations remain closer to the object region than the baselines on these examples. The quantitative counterpart of this observation is reported in Figure~\ref{fig:geo}.

\paragraph{\textbf{Conclusion of Qualitative Findings.}}
The visual evidence confirms that iBKD does not merely replicate the teacher's features but internalizes a more robust representational strategy. By generating semantically complete activations that are invariant to geometric shifts, iBKD proves to be a highly effective framework for training Vision Transformers in data-scarce and spatially dynamic environments.

\end{document}